\documentclass{article}

\DeclareUnicodeCharacter{2212}{-}

\usepackage{PRIMEarxiv}
\usepackage[utf8]{inputenc} % allow utf-8 input
\usepackage[T1]{fontenc}    % use 8-bit T1 fonts
\usepackage{hyperref}       % hyperlinks
\usepackage{url}            % simple URL typesetting
\usepackage{booktabs}       % professional-quality tables
\usepackage{amsfonts}       % blackboard math symbols
\usepackage{nicefrac}       % compact symbols for 1/2, etc.
\usepackage{microtype}      % microtypography
\usepackage{lipsum}
\usepackage{fancyhdr}       % header
\usepackage{graphicx}       % graphics
\graphicspath{{media/}}     % organize your images and other figures under media/ folder
\usepackage{amsmath}
\usepackage{amssymb}
\usepackage{booktabs}
\usepackage{epsfig}
\usepackage{svg}
\usepackage{placeins}
\usepackage{caption}
\usepackage{mathrsfs}

\title{CGA-PoseNet: Camera Pose Regression via \\a 1D-Up Approach to Conformal Geometric Algebra
\thanks{\textit{\underline{Citation}}: 
\textbf{Authors. Title. Pages.... DOI:000000/11111.}} 
}

\author{
  Alberto Pepe, Joan Lasenby \\
  Signal Processing and Communications Lab \\
  University of Cambridge \\
  Cambridge, UK\\
  \texttt{\{ap2219, jl221\}email@email} \\
}

\begin{document}
\maketitle

\begin{abstract}
We introduce CGA-PoseNet, which uses the 1D-Up approach to Conformal Geometric Algebra (CGA) to represent rotations and translations with a single mathematical object, the motor, for camera pose regression. We do so starting from PoseNet, which successfully predicts camera poses from small datasets of RGB frames. State-of-the-art methods, however, require expensive tuning to balance the orientational and translational components of the camera pose.This is usually done through complex, ad-hoc loss function to be minimized, and in some cases also requires 3D points as well as images. Our approach has the advantage of unifying the camera position and orientation through the motor. Consequently, the network searches for a single object which lives in a well-behaved 4D space with a Euclidean signature. This means that we can address the case of image-only datasets and work efficiently with a simple loss function, namely the mean squared error (MSE) between the predicted and ground truth motors. We show that it is possible to achieve high accuracy camera pose regression with a significantly simpler problem formulation. This 1D-Up approach to CGA can be employed to overcome the dichotomy between translational and orientational components in camera pose regression in a compact and elegant way. 
\end{abstract}

% keywords can be removed

\begin{figure}[h]
    \centering
    \includegraphics[width=\textwidth]{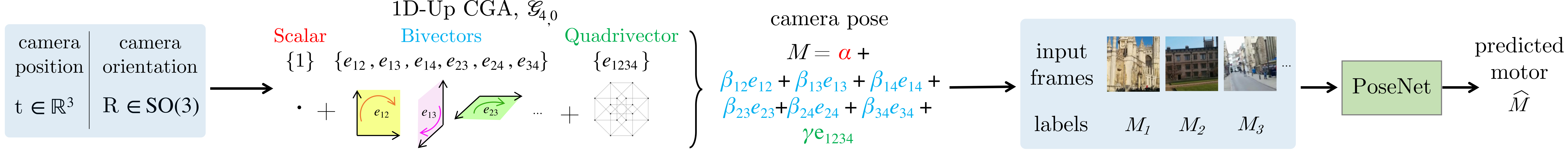}
    \caption{The CGA-PoseNet pipeline. A motor $M$ is an object in the 1D-Up Conformal Geometric Algebra with scalar, bivector and quadrivector components. A motor represents a translation $and$ a rotation: we represent camera poses as motors and trained PoseNet on them, showing how state-of-the-art accuracy in camera pose regression can be reached by simply minimizing the MSE loss between predicted and ground truth motors. There is no need to formulate a cumbersome and computationally expensive loss function that accounts for camera position and orientation separately as in previous literature.}
    \label{fig:my_label}
\end{figure}
\section{Introduction}

By camera pose regression we refer to the prediction of the camera position and orientation which is fundamental for many computer vision applications \cite{hoque2021comprehensive}, including augmented reality \cite{ababsa2004robust,sahu2021artificial}, robotics \cite{lee2020camera,yen2021inerf} and autonomous driving \cite{sauerbeck2022combined}. It has been historically solved via 2D-2D or 2D-3D feature matching methods, most famously the SIFT \cite{lowe2004distinctive}, SURF \cite{bay2008speeded} or ORB \cite{rublee2011orb} algorithms. Although very accurate, feature-based methods fail with texture-less objects, in cluttered scenes or in realistic weather conditions, and require a large database of features \cite{bergamo2013leveraging,wang2006coarse,zeisl2015camera}.

For this reason, several deep learning methods have been proposed in the literature. Despite not being as accurate as SIFT, deep neural networks can deal with smaller images and demonstrate high generalizability and faster inference times, even on small datasets \cite{en2018rpnet,melekhov2017relative,nakajima2017robust,rambach2016learning,yi2018learning}. No matter which architecture is employed, however, the orientational and translational components of the camera pose are generally still represented as two different mathematical objects which require  separate treatments \cite{chen2021wide,elmoogy2021pose,kendall2016modelling,kendall2017geometric,kendall2015posenet,xiang2017posecnn,xu2020estimating}.  This makes the pose regression task more expensive as it adds further tuning of the loss function on top of the hyperparameter tuning for a successful pose estimation. 

In this work, we reformulate the problem initially posed in \cite{kendall2015posenet} of learning the camera pose from RGB images by representing translation and rotation with a single mathematical object, the \textit{motor}. Motors sit in a 4D space with Euclidean signature, i.e. with all basis vectors squaring to $+1$ (see Figure 1). Our approach has three main advantages: \begin{itemize}
    \item A joint prediction of the camera position and orientation, unlike most previous literature which requires a tuneable weighting between the two. Note that we could have also used other elegant representations of rigid body transformations (screw theory/Pl{\"u}cker coordinates, CGA etc), but our 1D-Up approach to CGA provides a base space with Euclidean signature (therefore no inherent null structures)  which lends itself to the construction of simple loss functions. 
    
    \item The Euclidean signature of the space implies a simple, well-behaved loss function, namely the MSE between original and predicted motors. MSE is normally used in regression problems and easily computed. Loss functions presented in the literature, in fact, require an additional search over the weights for the orientational and translational components, which are dataset specific and add to the training complexity.
    
    \item Results comparable to several state-of-the-art methods \cite{kendall2016modelling,kendall2017geometric,kendall2015posenet,walch2017image} despite a simpler problem formulation and only using the camera pose as training label. No information on the 3D points of the scenes is employed at any point of our regression pipeline. 
\end{itemize}

{\bf Notation}. We will employ boldface lowercase letters for vectors (e.g. $\mathbf{t}$) and boldface uppercase letters for matrices (e.g. $\mathbf{R}$) when dealing with 3D Euclidean geometry. For Geometric Algebra (GA), on the other hand, we will stick to the notation commonly employed in the field \cite{doran2003geometric,lasenby20201d}, hence we will use simple lowercase letters for vectors (e.g. $e_1, e_2, a, t$, ...), uppercase letters for elements with grade 2 or higher (e.g. rotors $R, T_a, M$, ...) \textit{or} elements in CGA (e.g. $X = f(x), ...$) and Greek letters for scalars (e.g. $\lambda, \alpha, \gamma, ...$).

\section{Related Work}

{\bf Problem formulation}. We wish to estimate camera pose $\mathbf{p} \in \textrm{SE}(3)$, with $\textrm{SE}(3) \triangleq \{ (\mathbf{R}, \mathbf{t}) : \mathbf{R} \in \textrm{SO}(3), \mathbf{t} \in \mathbb{R}^3 \}$ via a deep neural network given an RGB image $I$ taken by a camera $C$. This is usually posed as a supervised learning problem in which the network predicts $\hat{\mathbf{p}}$ given the label $\mathbf{p} = [\mathbf{t}, \mathbf{q} ]$, with $\mathbf{t}$ being the camera position and $\mathbf{q}$ the camera orientation expressed as a quaternion \cite{kendall2017geometric,kendall2015posenet,xu2020estimating}. \newline

{\bf Choosing the loss function}. Much of the attention in camera pose regression problems has been focused on the loss function to be minimized rather than on the choice of representations for rotations and translations.

In \cite{kendall2015posenet}, the rotational and the translational part are empirically weighted together as follows

\begin{equation}
    \mathcal{L}_{\beta} = \mathcal{L}_{\mathbf{t}} + \beta \mathcal{L}_{\mathbf{q}} =  \| \hat{\mathbf{t}} -\mathbf{t} \|_2 + \beta \left\|\hat{\mathbf{q}} - \frac{\mathbf{q}}{\|\mathbf{q}\|}\right\|_2
\label{eq:loss1}
\end{equation} 

in which $\beta$ is a weighting scalar. However, the choice of $\beta$ is non trivial and a grid search is required. The optimal value was found to be ``the ratio between expected error of position and orientation at the end of training", which is not intuitive. Moreover, the value of $\beta$ varies significantly for each dataset, even if the volumes spanned by the datasets are comparable. For example, for the indoor datasets which are all $\leq 18\text{m}^3$, the optimal $\beta$ was found to be $\beta \in [120, 750]$. 

A similar range of values of $\beta$ is seen in Walch et al., who also used Equation \ref{eq:loss1} in \cite{walch2017image}: the pretrained GoogLeNet is followed by LSTM modules with two different fully connected layers, one for the position and one for the orientation, as last layers. 

The loss function of Equation \ref{eq:loss1} has also been employed in \cite{elmoogy2021pose}. In it, a pretrained ResNet50 convolutional neural network is used to extract features for each image, which are then reshaped into graph form and input to a graph neural network (GNN) to predict position and orientation. This more complex architecture allowed Elmoogy et al. to be less strict on the choice of $\beta$ and empirically fix $\beta = 10$ for indoor scenes and $\beta = 200$ for outdoor scenes.

A similar weighting has been proposed in \cite{xu2020estimating}. Xu et al. employed 2D trajectories of pedestrians to estimate the camera pose rather than RGB images only, and found that the weight parameter does not have a significant impact on the regression accuracy.

Weighting the translational and rotational parts is hence heavily dependent on the kind of datasets available and the chosen architecture.

The authors of \cite{kendall2015posenet} proposed a more advanced weighting strategy in \cite{kendall2017geometric}: \begin{equation}
     \mathcal{L}_{\sigma} = \mathcal{L}_{\mathbf{t}} \exp({−\hat{s}_{\mathbf{t}}}) +\hat{s}_{\mathbf{t}} + \mathcal{L}_{\mathbf{q}}\exp({-\hat{s}_{\mathbf{q}}}) +\hat{s}_{\mathbf{q}}
\label{eq:loss2}
\end{equation} with $\hat{s} := \log(\hat{\sigma}^2)$ being a learned weight and $\sigma^2$ the variance modelled through homoscedastic uncertainty. This probabilistic deep learning approach is superior to the $\beta$-weighting, but nonetheless still a weighting approach, with $\sigma_{\mathbf{q}}^2$ and $\sigma_\mathbf{\mathbf{t}}^2$ to be learned and possibly differing from each other by several orders of magnitude.

Also in \cite{kendall2017geometric}, a weighting-free approach is suggested: geometric reprojection error is used to combine the rotational and translational components into a single scalar loss. The geometric reprojection function $\pi$ is introduced, that maps a 3D point $\mathbf{g}$ to 2D image coordinates $(u,v)$: \begin{equation}
    \pi (\mathbf{t}, \mathbf{q}, \mathbf{g}) \mapsto (u, v)
\end{equation}where $\pi$ is defined via \begin{equation}
    (u', v', w') = \mathbf{K}(\mathbf{R}\mathbf{g} + \mathbf{t})
\end{equation} with $(u, v) = (u'/w', v'/w')$, $\mathbf{K}$ the intrinsic camera calibration matrix and $\mathbf{R}\in \textrm{SO}(3)$ the rotation matrix corresponding to $\mathbf{q}$. The proposed loss takes the norm of the reprojection error between the predicted and ground truth camera pose:

\begin{equation}
\mathcal{L}_{\mathbf{g}} = \frac{1}{|\mathcal{G}'|} \sum_{\mathbf{g}_i \in \mathcal{G}'}\|\pi(\mathbf{t}, \mathbf{q}, \mathbf{g}_i) - \pi ( \hat{\mathbf{t}}, \hat{\mathbf{q}}, \mathbf{g}_i ) \|_1
\label{eq:loss3}
\end{equation}

In which $\mathcal{G}'$ is a subset of all the points $\mathbf{g}$ in image $I$. Despite the high accuracy of this approach, the amount of computation required at each learning iteration is significantly higher than that required by Equations \ref{eq:loss1}-\ref{eq:loss2}. In addition, further discussion is needed to choose the most appropriate norm to be minimized. 

The dichotomy between rotational and positional components is also present in works that adopt completely different regression strategies. This is the case in Chen et al., who suggested an ad-hoc parameterization in \cite{chen2021wide}:  DirectionNet factorizes relative camera pose, specified by a 3D rotation and a translation direction, into a set of 3D direction vectors: the relative pose between two images, however, is still inferred in two steps for the rotation (DirectionNet-R) and the translation (DirectionNet-T) components. 

Works like \cite{chen2021wide,kendall2017geometric} show that efforts in unifying the rotational and translational components correspond to  better positional and rotational estimation. On the other hand, they are significantly less intuitive compared to the original PoseNet work \cite{kendall2015posenet}. In this work we wish to preserve the PoseNet pipeline, which is simple and successful,  but also to avoid the rotational and translational weighting. We do so through a new mathematical representation for camera pose. \newline 

{\bf Rotation representations}. While the position is most simply learned in Euclidean space, there are several possible ways to represent rotations. These include Euler angles, axis-angle form, rotation matrices, quaternions or, in GA, rotors and bivectors \cite{fang2018euler,huang2017deep,pavllo2018quaternet,pepe2021learning}.

It has been shown how different rotation representations might impact learning algorithms \cite{grassia1998practical,pepe2021learning,saxena2009learning,zhou2019continuity}. Euler angles, for example, suffer from gimbal lock, rotation matrices are over-parametrised and their orthogonality can be difficult to enforce in learning algorithms, and quaternions have a double mapping for each rotation. Already in \cite{grassia1998practical} the limitations of Euler angles and quaternions were highlighted when differentiation or integration operations were involved. Zhou et al., in \cite{zhou2019continuity} and Saxena et al. in \cite{saxena2009learning} both suggested that the discontinuity in the mapping from the rotation matrix $\mathbf{R} \in \textrm{SO}(3)$ onto a given representation space is responsible for large regression errors. In \cite{pepe2021learning}, the discontinuity issue is bypassed by representing rotations exclusively via rotors and bivectors in GA.

While most of the camera pose regression problems employ quaternions to represent rotation, the potential of GA in learning problems is still largely uncharted.

\section{Method}
The main idea of our paper is to represent camera poses with motors in a specific space. In this Section, we will provide the reader with the fundamental notions to understand the motor representation. We will first present basic definitions of GA in Section 3.1: grades, the geometric product, the rotor and the reversion operator. We will then add two additional basis vectors $e, \bar{e}$ to our space and map GA onto CGA in Section 3.2. In CGA, rigid body transformations are conveniently represented by rotors. Lastly, we will use $e$ as our origin to drop one dimension and work with a 1D-Up CGA space, which is a spherical space with curvature $\lambda$, in Section 3.3. 

\subsection{Geometric Algebra}
Starting from the second half of the last century, GA has found application in many fields, including physics \cite{doran2003geometric,hestenes2015space}, computer vision \cite{bayro2018geometric,hrdina2017binocular,wareham2004applications}, graphics \cite{hildenbrand2005geometric,hildenbrand2004geometric} and molecular modelling \cite{chys2008application,dorst2019boolean,lavor2019oriented}. 

GA is an algebra of geometric objects, which are built up via the {\em geometric product}. Given two GA vectors $a, b$, the geometric product is defined as \begin{equation}
    ab = a \cdot b + a \wedge b
\label{eq:geomprod}
\end{equation}

in which $\cdot$ represents the inner product and $\wedge$ represents the outer product. The outer product allows to define an object $A_r = a_1 \wedge a_2 \wedge ... \wedge a_r$, which we call an $r$-blade with {\it grade} $r$. Therefore, scalars are grade 0, vectors are grade 1, \textit{bivectors} are grade 2, \textit{trivectors} are grade 3, etc.  A linear combination of $r$-blades is called an $r$-vector. A linear combination of different $r$-vectors is a \textit{multivector}. In Equation \ref{eq:geomprod}, the inner product gives a scalar proportional to the cosine of the angle between $a,b$ and the outer product yields a bivector, which corresponds to the (signed) area of the parallelogram with sides $a,b$. Hence $ab$ is a \textit{multivector}, since it results from the sum of a scalar and a bivector, which have different grades: this is a substantial difference between geometric algebra and linear algebra.

Given a geometric product of vectors $R = a_1 a_2 ... a_r$ in an $n$-D space where $r\le n$, we define the reversion operator $\tilde{R} = a_r a_{r-1} ... a_1$. If we scale $R$ so that $R \tilde{R} = 1$ the expression \begin{equation}
R v \tilde{R}
\label{eq:rotor}
\end{equation} preserves both lengths and angles. This can be verified since $(R v \tilde{R})^2 = Rv \tilde{R} R v \tilde{R} = R v  v \tilde{R} = R v^2 \tilde{R} = v^2 R \tilde{R} = v^2$ and $(R v \tilde{R}) \cdot (R w \tilde{R}) = v \cdot w$. Hence, it can be shown that $R v \tilde{R}$ represents a rotation in GA. If $r$ is even, then $R$ is called a \textit{rotor}. ``Sandwiching" a GA object between a rotor and its reverse, as shown in Equation \ref{eq:rotor}, always results in a rotation of the object. 

A rotor is equivalent to a quaternion in 3D, it has fewer parameters than a rotation matrix, it does not suffer from  gimbal lock and can be extended to any arbitrary dimension. We will show how translations can also be represented via rotors in Section 3. We refer the reader to \cite{doran2003geometric} for a more complete discussion of GA fundamentals.

\subsection{Conformal Geometric Algebra (CGA)}

A geometric algebra with $p$ basis vectors that square to $+1$ and $q$ basis vectors that square to $-1$ is indicated via $\mathscr{G}_{p,q}$. CGA extends a geometric algebra by two additional basis vectors $e: e^2 = 1$ and $ \bar{e}: \bar{e}^2 = -1$, mapping $\mathscr{G}_{p,q}$ onto $\mathscr{G}_{p+1,q+1}$ \cite{wareham2004applications}. Hence, CGA maps the 3D geometric algebra $\mathscr{G}_{3,0}$, spanned by the basis vectors $\{e_1, e_2, e_3\}$, to $\mathscr{G}_{4,1}$, spanned by the basis vectors $\{e_1, e_2, e_3, e, \bar{e}\}$.

The two additional basis vectors $e, \bar{e}$ allow us to define two quantities: \begin{equation}
    n_{\infty} = e + \bar{e}
\end{equation} i.e. the infinity vector, and \begin{equation}
    n_0 = \frac{1}{2}(\bar{e} - e)
\end{equation}i.e. the origin vector. Points $x$ in a geometric algebra $\mathscr{G}_{p,q}$ are mapped to null vectors $X$ in $\mathscr{G}_{p+1,q+1}$ through $n_{\infty}$ and $n_0$ according to the equation
\begin{equation}
    X = f(x) = x + \frac{1}{2}x^2 n_{\infty} + n_0.
\end{equation} In CGA, reflections, rotations, translations and other geometric operations can all be conveniently represented as rotors. These rotors act on objects by sandwiching as in Equation \ref{eq:rotor}. In addition, CGA offers an intuitive representation of geometrical objects such point pairs, lines, planes, circles and spheres.   

\subsection{1D-Up CGA}
When dealing with transformations in Euclidean geometry in CGA, the point at infinity $n_{\infty}$ is kept constant. However, we can work with non-Euclidean geometries if we keep different quantities constant. For example by keeping $e$ constant, it can be shown that we are left with a \textit{hyperbolic} geometry. Similarly, when $\bar{e}$ is kept constant we are left with a \textit{spherical} geometry. Note how, by keeping either one of the bases $e$ or $\bar{e}$ constant, we have only $one$ additional basis vector compared to 3D GA (hence the name ``1D-Up") instead of two as in CGA (which is a ``2D-Up" space compared to 3D GA). \cite{lasenby2004recent,lasenby2011rigid,lasenby20201d}. 

In this paper we work with the latter case, in which $\bar{e}$ is kept constant and $e$ is our origin. The main advantages of the 1D-Up approach are: (1) a lower dimensionality of the space compared to CGA, (2) that the Euclidean nature of the space (i.e. that all basis vectors square to $+1$) allows us to construct a simple loss function that is invariant under rigid body transformations and (3) that both translations and rotations can be expressed as rotors, just as in the 2D Up space.

We now explain how to represent translations and rotations in a 1D-Up CGA. A translation by a 3D Euclidean vector $a$ has an equivalent rotor in 4D spherical geometry given by: \begin{equation} \begin{split}
    a \in \mathscr{G}_{3,0} \rightarrow T_a \in \mathscr{G}_{4,0}:\\
    T_a = g(a) = \frac{\lambda + ae}{\sqrt{\lambda^2 + a^2}}
    \label{eq:transvect}
    \end{split}
\end{equation} where $\lambda$ is a factor related to the radius of curvature of the spherical space. Note that translation by a vector $a$ is now  performed via a $rotor$ $T_a$, composed of a scalar part (with grade 0) and a bivector part (with grade 2). To translate an object in 1D-Up CGA we will ``sandwich" it between $T_a$ and $\tilde{T}_a$, as explained in Section 2.2.

It can be shown that a rotor $R$ in 3D Euclidean geometry is still $R$ in 4D spherical geometry. To rotate an object in 1D-Up CGA we will again ``sandwich" it between $R$ and $\tilde{R}$. 
The rigid body motion (i.e. translation and rotation) of an object $X$ into $X'$ in the 1D-Up CGA can hence be expressed as: \begin{equation}
    X' = T_a R X \tilde{R} \tilde{T_a} = M X \tilde{M}
\label{1duprot}
\end{equation}
in which $X$ is the corresponding representation in $\mathcal{G}_{4,0}$ of a point $x$  in $\mathcal{G}_{3,0}$, given by \begin{equation}
   X = h(x) =  \left(\frac{2\lambda}{\lambda^2 + x^2}\right) x + \left(\frac{\lambda^2 - x^2}{\lambda^2 + x^2}\right)e.
  \label{1dup}
\end{equation}
and in which we call the object $M = T_a R$ a \textit{motor}. The motor is obtained from the geometric product of $T_a$, the translation rotor, with $R$, the rotation rotor. $M$ is a multivector with a scalar part (grade 0), a bivector part (grade 2), and a quadrivector part (grade 4). It can hence be written as \begin{equation}
    M = \alpha +  \beta_{12} e_{12} + \beta_{13} e_{13} + \beta_{14} e_{14} + \beta_{23} e_{23} + \beta_{24} e_{24} + \beta_{34} e_{34} + \gamma e_{1234}
\end{equation}

with $e_{ij} = e_i \wedge e_j$, $e_{1234} = e_1 \wedge e_2 \wedge e_3 \wedge e_4$ and $e_4 \equiv e$. A 1D-Up CGA motor $M$ hence has a total of 8 coefficients $[\alpha, \beta_{12}, \beta_{13}, \beta_{14}, \beta_{23}, \beta_{24}, \beta_{34}, \gamma]$. Note that the motor is constrained to satisfy $M\tilde{M}=1$. 

 We can then represent the camera pose, i.e. its orientation (a rotation) $and$ its position (a translation), with the 8 components of a single motor. Because we are working in this spherical space with a Euclidean signature, it will be possible to construct loss functions on motors in this space which are well-behaved under minimisation.  
\section{Experiments}

A motor has a total of 8 coefficients to be predicted, one scalar, one quadrivector and 6 bivectors. This means 1 more parameter with respect to $[\mathbf{t}, \mathbf{q}]$, which are 3 + 4 = 7, but a single search space for our network to span. 

We employed the \textit{Cambridge Landmarks} \cite{kendall2015posenet} and the \textit{Scenes} \cite{shotton2013scene} datasets for our experiments. We recast the original labels  from $[\mathbf{t}, \mathbf{q}]$ for the \textit{Landmarks} datasets (or from $[\mathbf{t},\mathbf{R}]$, for the \textit{Scenes} dataset) into the motor coefficients $[\alpha, \beta_{12}, ... , \gamma]$. We work in a spherical space with curvature $\lambda$, where the rotation is the same as the rotation in Euclidean space and the translation uses the vector $\mathbf{t}$. This can be done easily as the rotor $R$ is isomorphic to the quaternion $\mathbf{q}$ in 3D, while the corresponding $T_t$ can be found from Equation \ref{eq:transvect}.  The pose motor is obtained as $M = T_t R$.  

Hence, CGA-PoseNet extends PoseNet to regress the camera’s pose expressed as a motor relative to a scene starting from a $224 \times 224$ RGB image. The original architecture Inception v3, derived from \cite{szegedy2015going,szegedy2016rethinking}, is pretrained on \textit{ImageNet}. We then: \begin{itemize}
    \item Substitute the training labels ([$\mathbf{t}, \mathbf{q}]$ or $[\mathbf{t},\mathbf{R}]$)  with motors $M$;
    \item Substitute the final softmax layer with a fully connected layer with dimensionality 8, to match the number of coefficients of the motor;
    \item Use the MSE loss between predicted and ground truth motors during training.
\end{itemize}  

\begin{table}[htbp]
\centering
\caption{The chosen curvature $\lambda$}
\resizebox{0.6\textwidth}{!}{
\begin{tabular}{ccccc}
\hline
\textbf{Dataset} & \textbf{Train} & \textbf{Test} &\textbf{Area / Volume} & \textbf{$\lambda$}  \\
\hline
Street  &  1500 & 1500 & 50000$\text{m}^2$ & 1000 \\
Great Court   & 1000 & 1000 & 8000$\text{m}^2$ & 200 \\
King's & 1220 & 343&  5600$\text{m}^2$&  200 \\
St. Mary's & 1487 & 530 & 4800$\text{m}^2$ &  200 \\
Old Hospital & 895 & 182 & 2000$\text{m}^2$& 200 \\

Shop & 231 & 103 &  875$\text{m}^2$ & 10 \\
RedKitchen & 3000 & 1000 & $ 18\text{m}^3$ & 10  \\
Office & 2000 & 2000 & $ 7.5\text{m}^3$ & 10  \\
7 Scenes& 1000 & 1000 & $\leq7.5\text{m}^3$ & 10  \\
\hline
\label{table:lambda}
\end{tabular}%
}
\end{table}

\begin{figure}[h]
\begin{center}
     \includegraphics[width=0.8\textwidth]{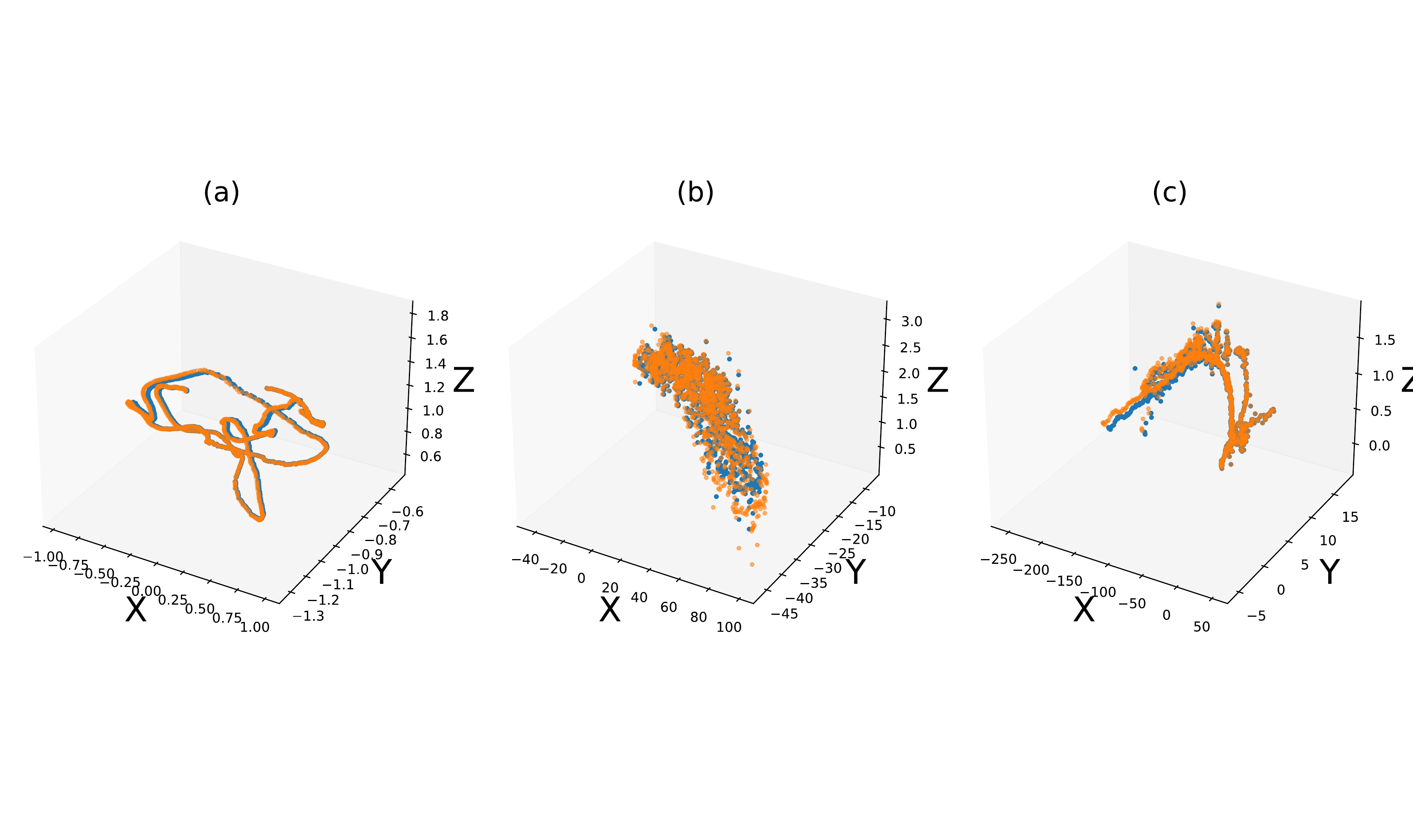}
   \caption{Camera trace in Euclidean (blue) and spherical geometry (orange) for the datasets: (a) $Office$, $\lambda = 10$, (b) $King's$, $\lambda = 200$, (c) $Street$, $\lambda = 1000$.}
\end{center}
\label{fig:lamb}
\end{figure}

{\bf Discussing $\lambda$, the curvature of the space}. $\lambda$ is the only free parameter in our approach and it controls the curvature of the 4D spherical space. In contrast to the weighing parameter $\beta$ of \cite{elmoogy2021pose,kendall2015posenet,xu2020estimating}, there is no need for a grid search, as we will see that it can simply be chosen to be directly proportional to the dataset area, before any training begins. Equivalently, by normalizing the input position data, it is possible to keep $\lambda$ fixed to a desired value. 

As $\lambda \rightarrow \infty$ the spherical space tends to the flat Euclidean space \cite{lasenby20201d}. For $\lambda \gg 1$, however, the motor coefficients containing $e_4$ are several orders of magnitude smaller than the others, negatively impacting the training.
On the other hand, a small $\lambda$ makes the curvature of the spherical space noticeable, meaning that our motor is not suitable to represent a roto-translation of a camera in the real world.

To choose an appropriate $\lambda$, we plotted the positional component associated with each frame (i.e. the camera trace) for every dataset. We compared the trace in Euclidean space (i.e. $\mathbf{t} = [x, y, z]$) with the trace $d$ in spherical space $\mathscr{G}_{4,0}$ as a function of $\lambda$. Examples are given in Figure \ref{fig:lamb}. The processing of extracting the trace in spherical space is explained in Section 4.3.

As we wish to represent a camera pose in 3D Euclidean space, we empirically chose $\lambda$ in such a way that (1) the curvature is not noticeable, i.e. $d \simeq \mathbf{t}$ and (2) the motor coefficients $[\alpha, \boldsymbol{\beta}, \gamma]$ are all within the same order of magnitude. For this reason, $\lambda$ can be picked to be proportional to the area spanned by the dataset. The choices of $\lambda$ for each of the datasets are summarized in Table \ref{table:lambda}.

\subsection{Datasets}

{\bf Cambridge Landmarks} \cite{kendall2015posenet} includes 6 datasets of outdoor scenes, namely \textit{Street, Great Court, King's College, St. Mary's Church, Old Hospital} and \textit{Shop Facade}, spanning areas ranging from $875\text{m}^2$ up to $50000\text{m}^2$. Each dataset includes a sequence of RGB frames with their corresponding labels, i.e. the motor coefficients $[\alpha, \beta_{12}, ... \gamma]$ representing the camera pose. Labels are generated via structure from motion, as described in \cite{wu2013towards}. We employed no more than 1500 frames for training. This is to show how, even with a smaller training set size, comparable results to the SoTA can be achieved. Each dataset presents significant clutter and the training and testing trajectories are non overlapping. \newline 
{\bf 7 Scenes} \cite{shotton2013scene} includes 7 datasets of indoor scenes, namely \textit{Office, Pumpkin, Red Kitchen, Heads, Fire, Stairs} and \textit{Chess}, each spanning a volume not larger than $18 \text{m}^3$. The dataset was intended for RGB-D relocalization and has been captured through a Kinect RGB-D sensor. We employed no more than 3000 frames for training. 

\subsection{Training details}

We train PoseNet in a supervised fashion by labeling each frame $I$ with the corresponding camera pose $M$. The loss we minimize is \begin{equation}
    \mathcal{L}_{M} = \textrm{MSE}(\hat{M},M)
    \label{eq:loss4}
\end{equation} where $\hat{M}$ and $M$ are the predicted and ground truth motors, respectively.
The network has been trained three times for each dataset with batch size $B = 64$, for $E = 100$ epochs and implementing early stopping with patience $P = 20$ during the last run to avoid overfitting. Results are measured with the weights obtained after the third training session. The optimizer has been kept to Adam with exponentially decaying learning rate, with initial value $\eta_0 = 10^{-4}$ and decay rate of $0.99$. The hyperparameters have been chosen empirically according to the chosen datasets, and combinations of $B = \{32, 64, 128\}$, $E = \{50, 100, 150, 200\}$, $P = \{8, 15, 20\}$, $\eta_0 = \{10^{-2}, 10^{-3}, 10^{-4}, 10^{-5}\}$ have also been tested but found to be suboptimal. The training takes an average of $580$ms per learning step, corresponding to about $8$s per epoch assuming the training set contains 1000 images. 

We leveraged transfer learning as discussed in \cite{kendall2015posenet} and employed the weights from $ImageNet$ (see \cite{deng2009imagenet}) for prior training  of CGA-PoseNet to ensure a successful regression even on small datasets such as those employed in this paper. 

All the experiments have been written as Jupyter notebooks on Google Colaboratory and run on a NVIDIA Tesla T4 GPU at 1.59 GHz. The Machine Learning architectures have been implemented via the Keras API of Tensorflow, 3D rotations have been handled through the Spatial Transformations package of Scipy and Geometric Algebra operations have been performed via the Clifford library \cite{python_clifford}. The Jupyter notebooks, predictions and measurements files are all included as Supplementary Materials.

\subsection{Metrics}
In order to evaluate the goodness of this regression strategy, we evaluate two metrics: (i) positional error and (ii) rotational error.

The positional and rotational error are computed by extracting the translational and rotational components from $\hat{M}$ and $M$. Given a motor $M$, this is done as follows: \begin{itemize}
    \item evaluate the displacement vector $D$ by transforming the origin, $e$, with $M$ (the origin is not affected by rotation so only the translation will have any effect), i.e. $ D = M e \tilde{M}$. The displacement vector has the form $D = \delta_1 e_1 + \delta_2 e_2 + \delta_3 e_3 + \delta_4 e$
    \item project $D \in \mathscr{G}_{4,0}$ onto $d \in \mathscr{G}_{3,0}$ via Eq. \ref{1dup} (i.e. $d = h^{-1}(D)$). From $d$ and $\hat{d}$ we compute the positional error
    \item evaluate the translation in spherical geometry $T_d = g(d) \in \mathscr{G}_{4,0}$
    \item retrieve the rotor as $R = \tilde{T}_d M$. From $R$ and $\hat{R}$ we compute the rotational error
\end{itemize} 
We will define positional error between original position $d$ and predicted position $\hat{d}$  as: \begin{equation}
        err_{pos} = \| \hat{d} - d\|_1\\
\end{equation} as this was found to perform the best with the chosen datasets and it does not increase quadratically with magnitude. In addition, it is in agreement with the norm chosen in the loss functions of \cite{kendall2017geometric} and Chapter 3 of \cite{kendall2019geometry}. 

The rotational error between original rotation $R$ and predicted rotation $\hat{R}$ has been inspired by the error to assess regressions on rotations in \cite{pepe2021learning,zhou2019continuity} and is defined as:
\begin{equation}
    err_{rot} = \cos^{-1}(\langle R \tilde{\hat{R}} \rangle_0)
\end{equation} where $\langle \cdot \rangle_0$ indicates the scalar part of the argument. Ideally, if $\hat{R} \simeq R$, then $R \tilde{\hat{R}} \simeq 1$. Hence, the rotational error is bounded in $[0, 180^{\circ}]$, which is what we would expect in rotation regression problems.

\section{Results}

\begin{table*}[t]
\centering
\caption{Median positional and rotational errors over the test set for 6 different approches. The last column shows the percentage of test frames with errors $<10m, <10^{\circ}$ obtained with CGA-PoseNet. }
\resizebox{\textwidth}{!}{%
\begin{tabular}{cccccccc}
\hline
\textbf{Scene} &\textbf{PoseNet \cite{kendall2015posenet}} & \textbf{\shortstack{Bayesian\\PoseNet} \cite{kendall2016modelling}} & \textbf{\shortstack{PoseNet\\LSTM} \cite{walch2017image}} & \textbf{\shortstack{PoseNet\\$\sigma^2$ Weights} \cite{kendall2017geometric}}& \textbf{\shortstack{PoseNet\\Geom. Repr.} \cite{kendall2017geometric}} & \textbf{\shortstack{CGA-\\PoseNet}} & $<10m, <10^{\circ}$ \\
\hline

Great Court  & - & - & - & 7.00m, 3.65$^{\circ}$ & 6.83m, \textbf{3.47}$^{\circ}$ & \textbf{3.77}m, 4.27$^{\circ}$ & 78.3$\%$\\

King's & 1.92m, 5.40$^{\circ}$ & 1.74m, 4.06$^{\circ}$& 0.99m, 3.65$^{\circ}$& 0.99m, 1.06$^{\circ}$& \textbf{0.88}m, \textbf{1.04}$^{\circ}$ & 1.36m, 1.85$^{\circ}$ & 95.6$\%$\\

Old Hospital    & 2.31m, 5.38$^{\circ}$ & 2.57m, 5.14$^{\circ}$&1.51m, 4.29$^{\circ}$ & 2.17m, 2.94$^{\circ}$ & 3.20m, 3.29$^{\circ}$& \textbf{2.52}m, \textbf{2.90}$^{\circ}$ &  99.4$\%$\\

St. Mary's  &2.65m, 8.48$^{\circ}$& 2.11m, 8.38$^{\circ}$&1.52m, 6.68$^{\circ}$ & \textbf{1.49}m, 3.43$^{\circ}$&1.57m, 3.32$^{\circ}$& 2.12m, \textbf{2.97}$^{\circ}$ & 87.5$\%$ \\

Shop &  1.46m, 8.08$^{\circ}$ &1.25m, 7.54$^{\circ}$& 1.18m, 7.44$^{\circ}$& 1.05m, 3.97$^{\circ}$&0.88m, \textbf{3.78}$^{\circ}$& \textbf{0.74}m, 5.84$^{\circ}$ & 89.3$\%$\\

Street  & - & - & - & 20.7m, 25.7$^{\circ}$ & 20.3m, 25.5$^{\circ}$ & \textbf{19.6}m,\textbf{19.9}$^{\circ}$ & 9.25$\%$\\
\hline
Chess & 0.32m, 6.60$^{\circ}$& 0.37m, 7.24$^{\circ}$& 0.24m, 5.77$^{\circ}$ & 0.24m, 5.77$^{\circ}$& \textbf{0.13}m, \textbf{4.48}$^{\circ}$& 0.26m, 6.34$^{\circ}$ & - \\
Pumpkin & 0.49m, 8.12$^{\circ}$& 0.61m, 7.08$^{\circ}$& 0.33m, 7.00$^{\circ}$& 0.25m, 4.82$^{\circ}$& 0.26m, \textbf{4.75}$^{\circ}$& \textbf{0.22}m, 5.18$^{\circ}$ & - \\
Fire & 0.47m, 14.0$^{\circ}$ & 0.43m, 13.7$^{\circ}$ & 0.34m, 11.9$^{\circ}$ & \textbf{0.27}m, 11.8$^{\circ}$ & \textbf{0.27}m, 11.3$^{\circ}$ & 0.28m, \textbf{10.3}$^{\circ}$ & - \\
Heads & 0.30m, 12.2$^{\circ}$ & 0.31m, 12.0$^{\circ}$ & 0.21m, 13.7$^{\circ}$ & 0.18m, 12.1$^{\circ}$ & \textbf{0.17}m, 13.0$^{\circ}$& \textbf{0.17}m, \textbf{7.98}$^{\circ}$ & - \\ 

Office & 0.48m, 7.24$^{\circ}$ & 0.48m, 8.04$^{\circ}$ & 0.30m, 8.08$^{\circ}$ & 0.20m, 5.77$^{\circ}$ & \textbf{0.19}m, \textbf{5.55}$^{\circ}$ & 0.26m, 7.23$^{\circ}$& -  \\

Red Kitchen & 0.58m, 8.34$^{\circ}$ & 0.58m, 7.54$^{\circ}$ & 0.37m, 8.83$^{\circ}$ & 0.24m, 5.52$^{\circ}$ & \textbf{0.23}m, \textbf{5.35}$^{\circ}$ &  0.55m, 16.7$^{\circ}$ & - \\

Stairs & 0.48m, 13.1$^{\circ}$ & 0.48m, 13.1$^{\circ}$& 0.40m, 13.7$^{\circ}$ & 0.37m, \textbf{10.6}$^{\circ}$ & 0.35m, 12.4$^{\circ}$ & \textbf{0.17}m, 12.0$^{\circ}$ & - \\
\hline
\label{table:posrot}
\end{tabular}%
}
\end{table*}

\begin{figure*}[ht]
\centering
   \includegraphics[width=\textwidth]{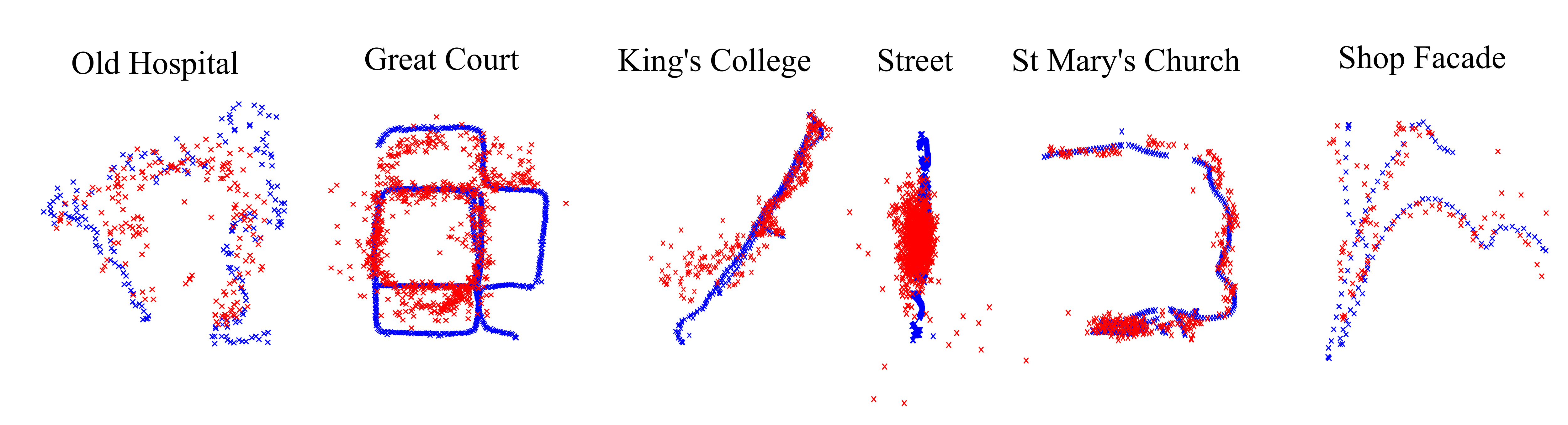}
   \caption{Aerial view of the predicted (red) and ground truth (blue) camera position $\hat{d}, d$ for the \textit{Cambridge Landmarks} test sets. Plots not to scale.}
\label{fig:position}
\end{figure*}

\begin{figure*}[hb]
\centering
   \includegraphics[width=\textwidth]{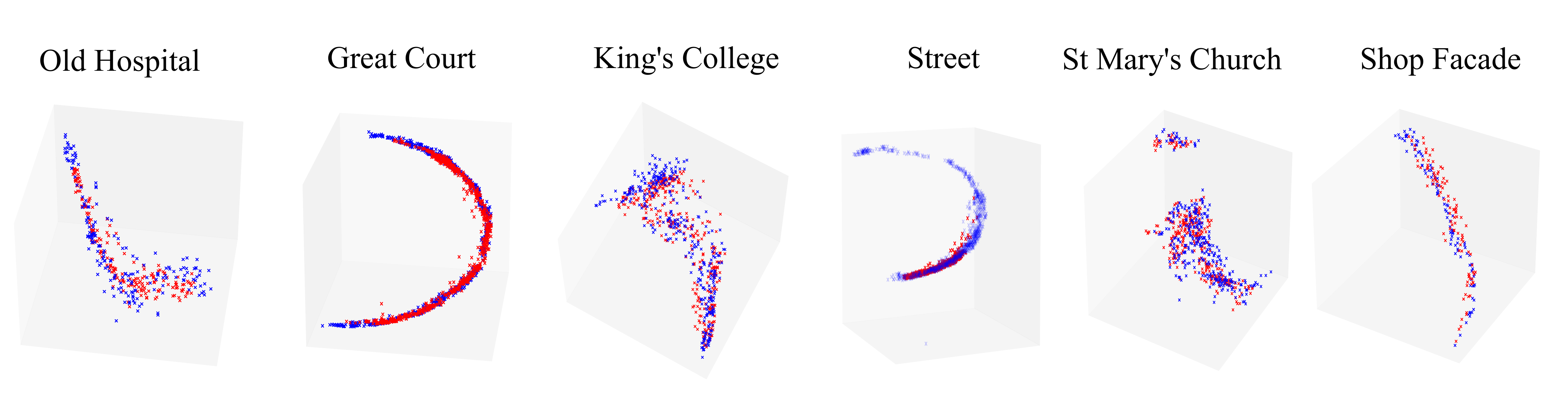}
   \caption{Predicted (red) and ground truth (blue) camera orientation (vector coefficients of $\hat{R}, R$) for the \textit{Cambridge Landmarks} test sets.}
\label{fig:orientation}
\end{figure*}

Results are summarized in Table \ref{table:posrot}.  We benchmarked our results with 5 different variations of the original PoseNet. Note how our CGA-PoseNet yields superior results in \cite{kendall2015posenet} and superior to comparable results in \cite{kendall2016modelling,kendall2017geometric,walch2017image}, even for larger and more difficult datasets like \textit{Street} and \textit{Great Court}. Specifically, our CGA-PoseNet approach is superior in every dataset analyzed to the results in \cite{kendall2015posenet}, and superior in 6 out of 13 datasets to \cite{kendall2017geometric} (with loss $\mathcal{L}_{\sigma}$) and in 7 out of 13 datasets (with loss $\mathcal{L}_{\mathbf{g}}$) despite having the simplest problem formulation of all of the compared approaches: our MSE loss, shown in Equation \ref{eq:loss4}, is significantly less computationally expensive than losses of Equations \ref{eq:loss1}, \ref{eq:loss2} and \ref{eq:loss3}, but still allowed us to regress the camera pose accurately. 

It is worth noting that our approach does not employ any additional information to the camera pose labels. For example, we completely discard the 3D points of the datasets on which the geometric reprojection error of \cite{kendall2017geometric} is based. The \textit{Cambridge Landmarks} dataset, for example, includes $> 2M$ 3D points, which are employed to formulate the loss $\mathcal{L}_{\mathbf{g}}$. Our approach yields comparable result to it without including any geometrical information of the scene during training, which makes CGA-PoseNet particularly suitable in scenarios in which points annotations are not available in the training set. 

Also note how our motor representation of the camera pose yields superior results to the approach in \cite{kendall2016modelling}, which includes the model uncertainty along with the 6 degrees-of-freedom representation of the camera pose, and generally superior results to the approach in \cite{szegedy2016rethinking}, which includes an additional LSTM network at the output of PoseNet.

\begin{figure*}[ht]
\centering
   \includegraphics[width=\textwidth]{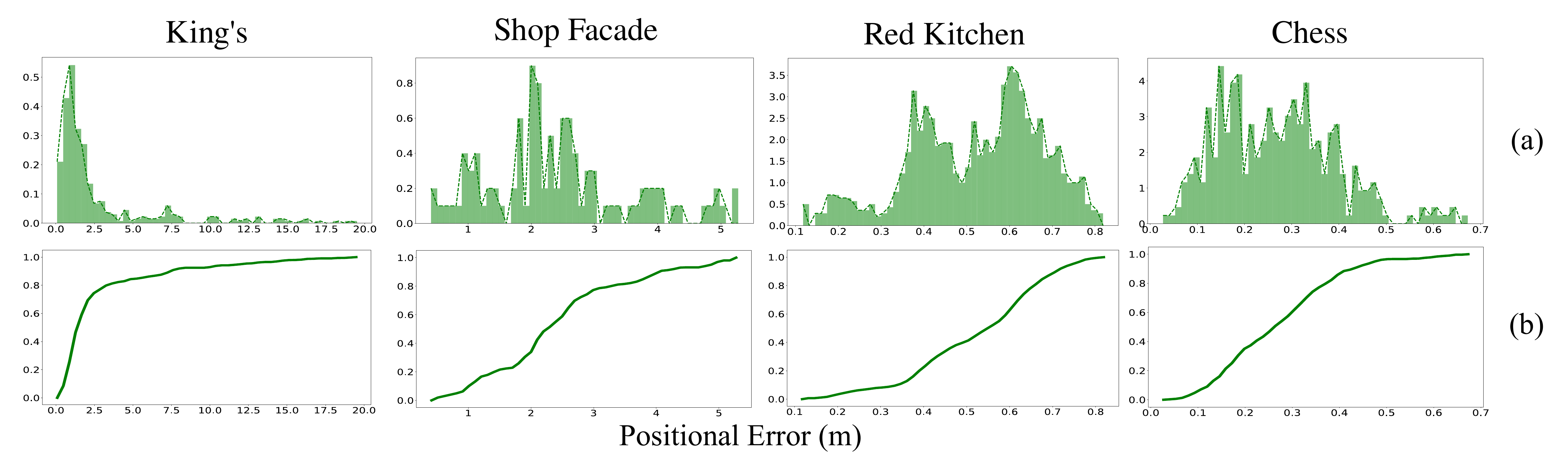}
   \caption{Positional error for selected datasets. (a) Normalized histogram and (b) cumulative distribution function over the test set. }
\label{fig:pos1}
\end{figure*}

\begin{figure*}[ht]
\centering
   \includegraphics[width=\textwidth]{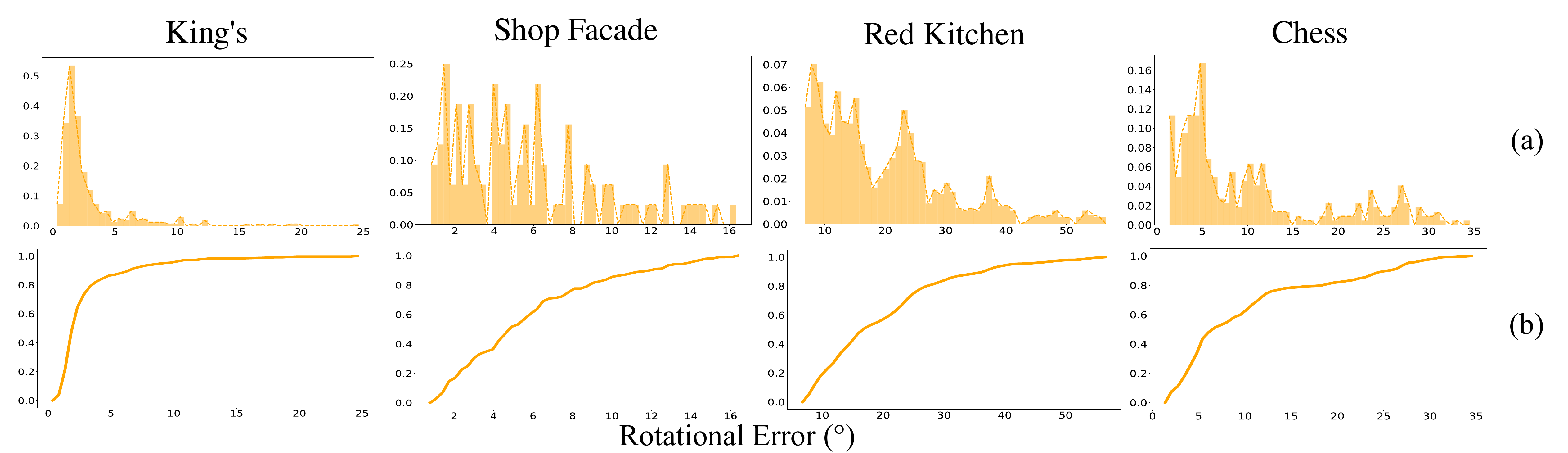}
   \caption{Rotational error for the selected datasets. (a) Normalized histogram and (b) cumulative distribution function over the test set. }
\label{fig:rot1}
\end{figure*}

\begin{figure}[htbp]
\centering
   \includegraphics[width=0.50\textwidth]{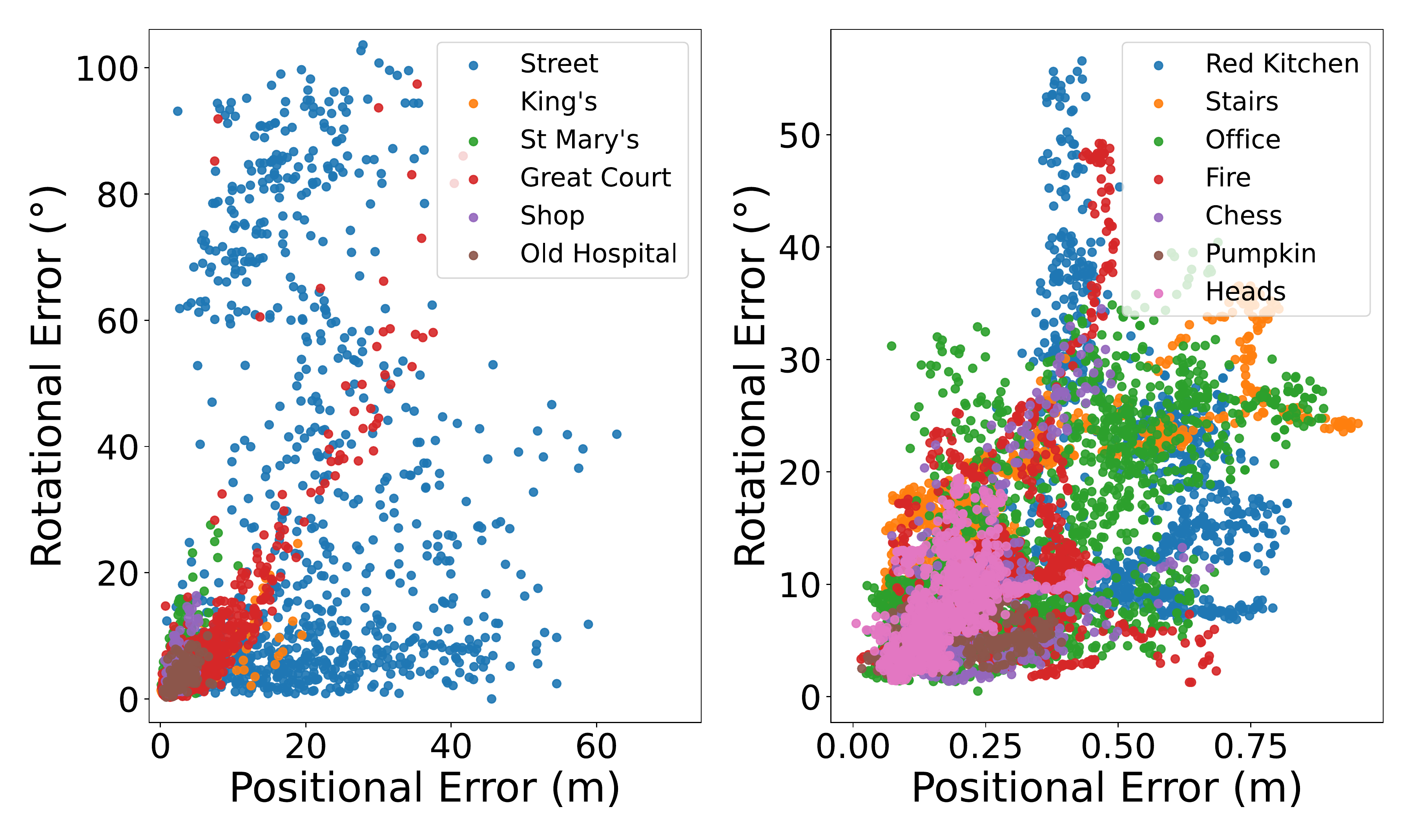}
   \caption{Error correlation for the two datasets.}
\label{fig:corr}
\end{figure}

Aerial views of the predicted and ground truth camera poses for the \textit{Cambridge Landmarks} are plotted in Figure \ref{fig:position}, as in \cite{kendall2015posenet}. The resulting traces represent the positions of the camera in 3D Euclidean space associated to the frames of the test set and corresponding predictions. If we consider only projections, the positional errors can be even lower than those in Table \ref{table:posrot}. The orientation has also been plotted as a 3D scatter plot of the vector coefficients of $\hat{R}, R$ (i.e. $e_{12}, e_{13}, e_{23}$) in Figure \ref{fig:orientation}.  As the positional and rotational errors are evaluated starting from the same quantity $M$, they are highly correlated (see Figure \ref{fig:corr}). The positional and rotational error distribution over the test sets of selected datasets are shown in Figs. \ref{fig:pos1} and \ref{fig:rot1}, respectively.

\begin{figure}[htbp]
\centering
   \includegraphics[width=\textwidth]{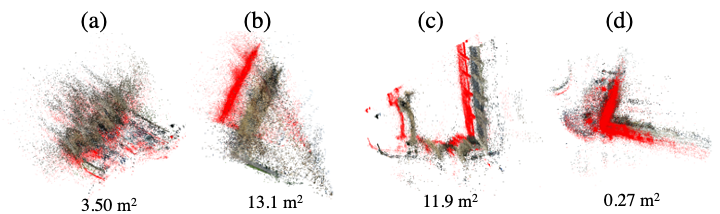}
   \caption{Point clouds with poses $M$ (colour) and $\hat{M}$ (red) and corresponding MSE between them. (a) \textit{King's}, (b) \textit{Old Hospital}, (c) \textit{St Mary's}, (d) \textit{Shop Facade}. The point clouds in the figure have been chosen with an MSE between them corresponding to the median MSE over the overall test sets.}
\label{fig:pointclouds}
\end{figure}

Lastly, we cross-checked the goodness of the pose predictions through point clouds. We generated the corresponding point clouds for four of the \textit{Cambridge Landmarks} set with smaller sizes, namely \textit{King's, St Mary's, Old Hospital} and \textit{Shop Facade}, from $\sim100$ frames through structure from motion \cite{agisoft2019agisoft}. We then measured the MSE between the point cloud with pose $M$ and the point cloud with pose $\hat{M}$ over the test sets of the four landmarks. The point clouds are plotted in Figure \ref{fig:pointclouds}: the point clouds in Euclidean space have been first projected onto the spherical space (via Eq. \ref{1dup}), then rotated and translated by $M$ and $\hat{M}$ (via Eq. \ref{1duprot}) and then projected back down to Euclidean space (via inverse Eq. \ref{1dup}), in which the MSE has been measured. 

This test tells us two things: first, that rotating and translating a point cloud in a spherical space with a suitable $\lambda$ is possible without noticeable deformation occurring, meaning that a motor $M$ in spherical space can be used to to give a good estimate of a camera pose in 3D Euclidean space; second, that the results obtained with point clouds are consistent with the positional errors of Table \ref{table:posrot}.

\section{Conclusions}

We have presented CGA-PoseNet, a GA approach for representing rotations and translations in the camera localization problem. To do this we used the 1D-Up approach to CGA, which unifies rotations and translations in a single object, the motor, which can be described by 8 (constrained) coefficients. We are also able to construct a well-behaved and simple cost function, thus eliminating the need for data dependent parameters. 

We have used CGA-PoseNet to learn the camera poses, expressed as motors, of 13 datasets , and demonstrated that this approach allows a much simpler formulation of the loss function while giving results comparable, when not superior, to more complex approaches existing in the literature. Results obtained with CGA-PoseNet, moreover, are comparable to those obtained minimizing the geometric reprojection error, which requires the 3D points in the model in order to be trained, which means we can achieve similar results with much less information as input to our model.

This paper aims to present a {\em proof of concept}  for a unifying approach to dealing with rotations and translations in camera pose problems. There is room for improvement in the results presented here -- a more rigourous tuning of the training hyperparameters could potentially provide results surpassing state-of-the-art methods.

%Bibliography
\bibliographystyle{unsrt}  
\bibliography{references}

\end{document}